\documentclass[10pt,a4paper,twocolumn]{article}
\usepackage[T1]{fontenc}
\usepackage[utf8]{inputenc}
\usepackage{lmodern}
\usepackage[a4paper,top=2.2cm,bottom=2.5cm,left=1.8cm,right=1.8cm,columnsep=0.65cm]{geometry}
\usepackage{microtype}
\usepackage{graphicx}
\usepackage{color}
\usepackage{algorithm}
\usepackage{algorithmic}
\usepackage{amsmath}
\usepackage{amssymb}
\usepackage{tikz}
\usetikzlibrary{fit,positioning,arrows.meta}
\usepackage[font=small,labelfont=bf,labelsep=period]{caption}
\usepackage[
    colorlinks=true,
    citecolor=blue,
    linkcolor=blue,
    urlcolor=blue
]{hyperref}
\usepackage{orcidlink}

\providecommand{\doi}[1]{\href{https://doi.org/\detokenize{#1}}{\url{https://doi.org/#1}}}

\newcommand{\paperTitle}{Analysing the Narrative Paradox through Formal Models and Properties in Interactive Narrative}
\hypersetup{
    pdftitle={\paperTitle},
    pdfauthor={Jules Clerc, Mohamed Sallak, Domitile Lourdeaux, Marc Ravaine},
    pdfkeywords={Experience Management, Narrative Paradox, Formal Description}
}

\newif\ifamnotice
\amnoticefalse
\newcommand{\amBookTitle}{[book title]}
\newcommand{\amYear}{[year]}
\newcommand{\amDOI}{[DOI]}

\makeatletter
\renewcommand\section{\@startsection{section}{1}{\z@}%
    {-2.6ex \@plus -1ex \@minus -.2ex}{1.3ex \@plus .2ex}%
    {\normalfont\large\bfseries}}
\renewcommand\subsection{\@startsection{subsection}{2}{\z@}%
    {-2.2ex \@plus -.8ex \@minus -.2ex}{1ex \@plus .2ex}%
    {\normalfont\normalsize\bfseries}}
\renewcommand\subsubsection{\@startsection{subsubsection}{3}{\z@}%
    {-1.6ex \@plus -.5ex \@minus -.2ex}{-0.5em \@plus -0.22em \@minus -0.1em}%
    {\normalfont\normalsize\bfseries}}
\makeatother
\begin{document}

% ---------------------------------------------------------------------
%  Title block (spans both columns)
% ---------------------------------------------------------------------
\twocolumn[{%
\begin{center}
    {\LARGE\bfseries \paperTitle\par}
    \vspace{1.1em}
    {\large
    Jules Clerc\textsuperscript{1,2}\,\orcidlink{0009-0009-0825-4498},\quad
    Mohamed Sallak\textsuperscript{1}\,\orcidlink{0000-0003-4803-2362},\quad
    Domitile Lourdeaux\textsuperscript{1}\,\orcidlink{0000-0002-3354-7294},\quad
    Marc Ravaine\textsuperscript{2}\,\orcidlink{0009-0008-7939-1740}\par}
    \vspace{0.7em}
    {\small
    \textsuperscript{1}Université de technologie de Compiègne, CNRS, Alliance Sorbonne Université, Heudiasyc, Compiègne, France\\
    \textsuperscript{2}ALTEN, Sophia Antipolis, France\\[0.3em]
    \texttt{\{jules.clerc,mohamed.sallak,domitile.lourdeaux\}@hds.utc.fr}\par}
\end{center}
\vspace{0.4em}
\begin{center}
\begin{minipage}{0.9\textwidth}
\small
\noindent\textbf{Abstract.}
Interactive narratives aim to let players influence the unfolding of a story while preserving a coherent and satisfying narrative progression. This creates a central tension, commonly described as the narrative paradox, between player autonomy and narrative control. Experience Managers (EMs) manage the story by determining how the narrative progresses in response to player behaviour. Although this paradox is widely discussed, it is rarely modelled in a way that supports systematic analysis, simulation, and comparison across management strategies. This paper proposes a formal framework for modelling and analysing the narrative paradox as a trade-off between interdependent properties shaped by EM decisions. We represent interactive narrative behaviour through an extended state-transition system in which world states are connected by player actions and system events. From this model, we formalize key properties involved in the paradox such as the player agency, EM controllability, author intent, and robustness, and define quantitative indicators derived from simulated plans. These indicators make it possible to compare EM strategies according to their ability to balance freedom, control, and narrative satisfaction. A case study compares different EM policies, showing how simulations can reveal correlations and conflicts between properties and identify compromise zones in the design space. This work contributes a reusable modelling and analysis framework for studying the narrative paradox and evaluating heterogeneous interactive narratives on shared formal grounds, with a focus on supporting the comparison and testing of EM strategies.

\end{minipage}
\end{center}
\vspace{1.2em}
}]

\ifamnotice
\begingroup
\let\thefootnote\relax
\footnotetext{\scriptsize This is an Accepted Manuscript version of the following work: J.~Clerc, M.~Sallak, D.~Lourdeaux, M.~Ravaine, \textit{\paperTitle}, in \amBookTitle, \amYear, Springer. This version of the manuscript has been accepted for publication, after final editorial and peer review (where applicable) is complete, but is not the Version of Record and does not reflect post-acceptance improvements (such as copyediting or typesetting), or any corrections. The final authenticated version is available online at: \url{https://doi.org/\amDOI}. Use of this Accepted Manuscript version is subject to the publisher's Accepted Manuscript terms of use: \url{https://www.springernature.com/gp/open-research/policies/accepted-manuscript-terms}}
\endgroup
\fi
\section{Introduction}
% 1) Définition d'un SNI et du rôle de l'EM et autres acteurs.
Interactive Narrative Systems (INS) define the overall framework in which an interactive story unfolds, including the narrative world, player actions, system events, goals, and possible story trajectories. Within this framework, the Experience Manager (EM) is the decision-making component responsible for managing story progression. Interactive narratives therefore seek to reconcile two seemingly contradictory objectives: offering players freedom of action while preserving a coherent and satisfying narrative experience. To address this tension, many INS rely on EM mechanisms that guide the narrative by intervening in the world according to the user's actions \cite{riedlInteractiveNarrativeIntelligent2013}.

% 2) évaluer et comparer ces systèmes reste difficile
Despite advances, evaluating and comparing these systems remains difficult. On one hand, the literature relies on a variety of representations \cite{thueUnifiedUnderstandingExperience2018}. This reflects the diversity of stories and forms of interactivity, but there is no commonly used layer for comparing the behaviours produced by heterogeneous EMs. On the other hand, the properties to be evaluated cover concepts such as player agency, EM controllability, robustness, or variability, but their definitions and measurement methods vary from one study to another.

% 3) Nous proposons ceci

We propose a formal modelling and analysis framework for studying the narrative paradox as a set of measurable trade-offs between INS properties, with the specific aim of comparing and testing EM strategies under a shared analytical representation. We rely on a formalism that models an INS as a state transition system where the player proposes actions and the EM chooses the effective transitions. The plans used for our analysis are generated by simulation, notably from a probabilistic model of which action a player is likely to propose in a given state. This does not change the representation of states and transitions.

In contrast with prior work aimed at designing or implementing an EM (e.g. Open-Theatre \cite{xuOpenTheatreOpenSourceToolkit2025} or IBSEN \cite{han-etal-2024-ibsen}), our objective is to introduce a formal model that evaluates the results produced by heterogeneous approaches, whether formal (planning-based problems) or generative (LLMs), as long as they expose an explicit world state. Rather than replacing the formalisms used internally by EM designers, our model acts as an "intermediate representation": a common structural layer of states, transitions, and plans onto which the output of an EM can be mapped and compared. We therefore require designers to map their EM's output behaviour onto this formalism, making evaluation interoperable and providing the explicit structural rigour often lacking in recent INS implementations, particularly LLM-based ones \cite{han-etal-2024-ibsen}.

From this model, we introduce a three-step methodology: (1) defining and analysing the use of a property in the state of the art, (2) translating this property into a formal specification, and (3) proposing evaluation indicators.

% 4) Nous proposons aussi une étude de cas
Finally, we illustrate the framework through a case study on the narrative paradox. Its goal is not to fully validate the proposed indicators, but to show how the formal model operates and how the properties are used in practice. Keeping the narrative design fixed and varying only the EM policy, we show how different management strategies shift the balance between properties during simulations. The results show the paradox as a compromise, where improving some properties degrades others, which the framework makes explicit and analysable.

\section{Related Work}
INS are studied through the tension between player freedom and narrative control, often referred to as the narrative paradox \cite{aylettEmergentNarrativeSocial2000,rothLudonarrativeHermeneuticsWay2018}. Experience Management has been proposed as a way to regulate this tension by adapting, constraining, or guiding the narrative according to the player's behaviour and the system's objectives \cite{riedlInteractiveNarrativeIntelligent2013}. However, comparing EM approaches remains difficult because they often rely on different assumptions, representations, and evaluation criteria.

A useful distinction can be made between the design of an INS and its representation. The design corresponds to the mechanisms used to produce the interactive experience, while the representation defines the concepts and structures used to describe and analyse the system \cite{thueUnifiedUnderstandingExperience2018}. Since representations are often chosen according to the system's design, the literature contains many distinct formalisms, such as state machines \cite{coteTextWorldLearningEnvironment2019a}, planning problems \cite{porteousApplyingPlanningInteractive2010,rivera2024story}, and graph-based models \cite{riedlLinearStoryGeneration2006c}. Each is well suited to the design it serves, but they are generally evaluated individually and rarely compared on shared grounds \cite{magerkoComparativeAnalysisStory2007}. Recent LLM-based systems sharpen the difficulty: their generative capabilities are promising, but their outputs remain hard to evaluate systematically \cite{koenitzCanAICreate2025}.

INS can be analysed from both structural and functional perspectives. Structural analysis focuses on the architecture of the system, its modules, and the organisation of the story space \cite{moriStructuredAnalysisExperience2019}. Functional analysis focuses instead on the behaviour produced during execution: the trajectories generated, the player's actions, and the resulting narrative experience. Such evaluations can rely on user studies, questionnaires or logs \cite{szilasObjectiveMetricsInteractive2014}. We focus on functional analysis through simulated plans to compare the behaviours produced by different EM policies.

Evaluation criteria are referred to in the literature as desiderata, dimensions, or properties \cite{robertsSurveyQualitativeAnalysis2008,riedlInteractiveNarrativeIntelligent2013,szilasObjectiveMetricsInteractive2014}. Among them, player agency describes the player's ability to act meaningfully and influence the story \cite{riedlGameAIStorytelling2011,cardona-riveraForeseeingMeaningfulChoices2014}, while EM controllability refers to the system's ability to influence or constrain narrative progression through events, constraints, pruning, or repair mechanisms \cite{nelsonSearchbasedDramaManagement2005,porteousApplyingPlanningInteractive2010,wareMultiAgentNarrativeExperience2019}. Other properties, such as author intent and robustness, are also central to maintaining a satisfying narrative structure: author intent captures alignment with the author's expected narrative trajectory, while robustness captures the system's ability to avoid blocked or problematic states \cite{zhuAuthorIntentionBased2012,riedlRobustAuthorableMultiplayer2011}.

\section{Formal Framework}
\subsection{Background}
% TODO : Voir par rapport à ce que l'on veut faire dans l'article si c'est pertinent d'avoir un système déterministe ou non-déterministe.
% TODO : Parler du manque de formalisme et ce que cela engendre.
% Dans cette section on peut parler du formalisme de l'article précédent et l'analyse fonctionnelle.

To support the functional comparison of behaviours produced by heterogeneous EMs, we introduce the common analytical vocabulary used throughout this paper. We use a formalism in the form of an extended-state machine \cite{clercModelingInteractiveNarrative} where:

Let $\mathfrak{S} = (S, T , \gamma, s_{\text{init}}, S_{\text{goal}})$ be a state transition system where :

\begin{itemize}
    \item $S = \{s_1, s_2, \dots\}$ is a finite set of states.
    \item $T = A \cup E$ (with $A \cap E = \emptyset$) is a finite set of transitions (where $\emptyset$ represents the empty set). Since we consider the EM to be the system-side actor, we partition T into two sets : $A$ for transitions that the player can use and $E$ for transitions that only the EM can instantiate.
    \begin{itemize}
        \item $A = \{a_1, a_2, \dots\}$ is a finite set of actions.
        \item $E = \{e_1, e_2, \dots\}$ is a finite set of events.
    \end{itemize}
    \item $\gamma : S \times T \to S$ is a state transition function. We note the domain $dom(\gamma) \subseteq (S \times T)$ such that :
    \begin{itemize}
        \item $\gamma(s,t) \text{ is defined } \Leftrightarrow (s,t) \in dom(\gamma)$
        \item $\gamma(s,t) \text{ is not defined } \Leftrightarrow (s,t) \notin dom(\gamma)$ 
    \end{itemize}
    \item $s_{\text{init}} \in S$ is the initial state of the system.
    \item $S_{\text{goal}} \subseteq S$ is the set of final goal states of the system.
\end{itemize}

The system is finite but not fixed: some EM policies extend it during execution by adding states, transitions, or entries in the transition function.

This system represents the story designed by the \textit{Author}. The \textit{Experience Manager} can develop and modify this system, while the \textit{Player} can suggest actions to instantiate. The objective is for the player to reach a target state $s_g \in S_{\text{goal}}$ from an initial state $s_{\text{init}}$. The sequence of states traversed by the system $\mathfrak{S}$ during a session is called the plan, denoted by $\pi$.

% Expliquer pourquoi on choisit cette représentation
We chose this representation because it combines readability and rigour: it supports graphical visualisation as a story graph understandable to authors, while providing a formal framework for defining properties, simulating, and measuring results. It is also open to new properties and indicators, and its closeness to graphs and state machines eases interoperability. The size of this state space can grow quickly with the scale of the narrative. This does not affect the analyses proposed here because our indicators are computed over the plans actually traversed during simulation, and never require the state space to be enumerated. We do not claim that every design can be mapped onto this representation: it assumes an explicit world state at each discrete time step. Planning-based designs satisfy this by construction, which explains their presence in our case study. Systems that produce the narrative directly as text, such as AI Dungeon \cite{latitudeAIDungeon}, maintain no comparable state: mapping them would require reconstructing one from the generated text.

\subsection{Property Definition}

% Introduction de la section
We will focus on defining a property and formalising it to (1) avoid any confusion that may arise from subjective definitions, (2) enable the improvement of these properties through explicit formal representation. Our methodology for defining a property consists of three steps.

The first step reviews how each property is addressed in the state of the art. We present and discuss existing definitions, which may differ on specific points, to clarify our design objectives and resolve ambiguities.

The second step translates these definitions into formal specifications based on the model presented above.

Thirdly, we design indicators to assess the property in practice, as quantitative measures obtained by simulation or by testing campaigns on human players. This links the formal definition to concrete issues such as comparing INS, adjusting EM parameters, or exploring trade-offs between conflicting properties. Our indicators are intentionally simple and directly calculable from simulated plans, providing operational measures for analysing and comparing EM configurations.

\section{Property modelling}
We apply this methodology to a subset of properties directly related to the narrative paradox: first the paradox itself, then the properties required to analyse the underlying trade-off. These formalisations describe what can be observed in the plans a system produces, not what players experience. Several of the properties modelled below are also used in the literature to designate experiential notions, and we do not claim that our definitions capture that dimension, which would require validation with human participants.

\subsection{The Narrative Paradox}
\subsubsection{Definition}

The narrative paradox refers to the tension between player agency and EM control. On one hand, players want to interact freely with the environment, make decisions and influence the story. On the other, the EM wants to control the narrative because it seeks to ensure a coherent and satisfying narrative structure \cite{aylettEmergentNarrativeSocial2000,rothLudonarrativeHermeneuticsWay2018}.

What certain authors mean by "satisfying structure" is often unclear when defining the narrative paradox. For the purpose of this work, we adopt a deliberately simplified and operational definition, tailored to our analysis needs. We consider a narrative structure to be satisfying if the player does not remain stuck in a problematic state, reaches an objective state, and progresses through the key objectives set by the author. This definition allows us to formalise and quantify satisfaction within our framework, although it does not aim to capture the full richness of narrative experience.

\subsubsection{Formalism}

To study the tension between player agency and EM controllability, we define three parameters in $[0,1]$: $scale_{agency}$ for the player's freedom to interact ($0$: no freedom, $1$: absolute freedom); $scale_{control}$ for the control the EM exercises over the story ($0$: no control, $1$: total control); and $scale_{satisfying}$ for the quality of the generated story, defined as the "satisfying structure" ($0$: very unsatisfying, $1$: very satisfying).

The narrative paradox therefore consists of finding a compromise between the freedom given to the player ($scale_{agency}$) and the control exercised by the EM ($scale_{control}$) to optimise the quality of the story ($scale_{satisfying}$). Framing this relationship as a strict trade-off warrants closer examination. Prior work indicates that perceived agency and narrative structure tend to be positively correlated in human-led sessions \cite{fisherStructureAgencyImprovisation2025}, which motivates the use of our indicators to identify EM configurations that reinforce both properties rather than treating them as mutually exclusive.

\subsubsection{Associated indicators}

We define a function $f(scale_{agency},scale_{control})=scale_{satisfying}$ giving the quality of the story generated by the INS for a plan $\pi$ or a set of plans $\Pi$. Each scale is instantiated using simple plan-based metrics, providing an interpretable and reproducible approximation of the underlying properties. The goal is then to find the values of $scale_{agency}$ and $scale_{control}$ that maximise $f$:

\begin{equation}
    \resizebox{.91\linewidth}{!}{$
            \displaystyle
            \underset{scale_{agency},\ scale_{control} \in [0,1]}{\operatorname{argmax}} \; f(scale_{agency}, scale_{control})
        $}
\end{equation}

Where $\operatorname{argmax}$ denotes the set of maximisers of $f$ over the domain. To model the narrative paradox, we need to define indicators related to Player Agency and EM controllability. Regarding the satisfaction scale of the generated narrative structure, we will then model the author's intention (to see if the player goes through the key objectives) and robustness (to see if the system can avoid situations that block the narrative's progress).

\subsection{EM controllability}
\subsubsection{Definition}

Control refers to an actor's ability to influence or constrain the unfolding of the story. For the player, this refers in particular to player agency, which allows them to choose actions to direct the narrative. The EM has greater control over the various elements of $\mathfrak{S}$.

\subsubsection{Formalism}

We define controllability as the actor's ability to modify each element of the system $\mathfrak{S}$. For example, an LLM-driven INS can dynamically add new states to the initial set $S$ according to the player's actions \cite{pengPlayerDrivenEmergenceLLMDriven2024}. In graph-based approaches, control may be achieved by pruning the narrative graph, for example by removing transitions, and evaluated through indicators such as the pruning rate \cite{wareMultiAgentNarrativeExperience2019}. Control can thus add, delete or modify states and transitions. We focus on the EM's choice of the transitions applied through $\gamma$.

The EM is responsible for triggering transitions within the system. It determines
the strategy to be implemented based on the action proposed by the player, denoted
$a_{prop} \in A \cup \{\perp\}$, where $\perp$ denotes the absence of a proposal. We define the function $\mu_{\text{EM}}(s,a_{prop}) = t_{select}$ to model the EM transition selection process $t_{select}$, such that $\mu_{\text{EM}} : S \times (A \cup \{\perp\}) \to T_s \text{ where } T_s = \{ t \in T \mid (s,t) \in dom(\gamma) \}$.

\subsubsection{Associated indicators}

Inspired by SBDM \cite{nelsonSearchbasedDramaManagement2005}, which proposes different types of events for managers to influence the progression of the narrative, we measure the EM's control as the share of events among the transitions performed. For a plan $\pi = (s_0,s_1,...,s_n)$ and the associated sequence of transitions performed $T(\pi) = (t_0,t_1,...,t_{n-1})$ where $s_{i+1} = \gamma(s_i,t_i)$, we define the number of transitions performed $N_{t}(\pi) = |T(\pi)|$ and the number of events performed $N_{e}(\pi) = \sum^{n-1}_{i = 0} [t_i \in E]$. Thus, the event rate for a plan $\pi$ is :

\begin{small}
    \begin{equation}
        EventRate(\pi) = \frac{N_{e}(\pi)}{N_{t}(\pi)}
    \end{equation}
\end{small}

For a set of plans $\Pi$, we average over plans:

\begin{small}
    \begin{equation}
        EventRate(\Pi) = \frac{1}{|\Pi|} \sum_{\pi \in \Pi} EventRate(\pi)
    \end{equation}
\end{small}

\subsection{Player agency}
\subsubsection{Definition}

Player agency is described in the literature both as the player's ability to do whatever they want at any time \cite{riedlGameAIStorytelling2011}, and as the satisfying power to take meaningful actions and see the results of their decisions and choices \cite{cardona-riveraForeseeingMeaningfulChoices2014}. The first formulation concerns what the system allows the player to do, and is readable from the transitions of a plan. The second concerns how players perceive their own influence, and can only be established with human participants. We formalise the first.

\subsubsection{Formalism}

We characterise agency through the player's ability to act. To give the player freedom, they must be able to act in non-terminal states. We denote by $S_{end}$ the set of terminal states. Formally, $\forall s \in S \backslash S_{end}, \ \exists a \in A$ such that $(s,a) \in dom(\gamma)$, which guarantees at least one available action in each of these states. Systems differ in when this is allowed: some let the player act at any time (\textit{Anytime}), while others only allow it at prescribed moments (\textit{Specific Time}) \cite{luoReviewInteractiveNarrative2015}. They also differ in how an action is entered, either as a free-text command or as a selection from a presented list. In all cases, these input modes reduce to selecting a transition outgoing from the current state.

\subsubsection{Associated indicators}
Among the metrics defined on player actions and system responses~\cite{mateasFacadeExperimentBuilding2003a,szilasObjectiveMetricsInteractive2014}, which can be computed over generated plans, we use the acceptance rate of the actions the player proposes to the EM, capturing the freedom given to the player.

For a plan $\pi = (s_0,...,s_n)$, let $A_{prop} = (a_0,...,a_{n-1})$ be the player's proposals, each $a_i \in A \cup \{ \perp \}$, and let the EM apply $\mu_{EM}(s_i,a_i) = t_i \in T$.

We count the player's proposals $N_{prop}(\pi,A_{prop}) = \sum_{i=0}^{n-1} [ a_i \in A ]$, and those accepted by the EM $N_{acc}(\pi,A_{prop}) = \sum_{i=0}^{n-1} [ a_i \in A \wedge \mu_{EM}(s_i, a_i) = a_i ]$.

The indicator is:

\begin{small}
    \begin{equation}
        % HAL version: split over two lines to fit the column width
        \begin{split}
        AcceptRate(\pi,A_{prop}) &= \frac{N_{acc}(\pi,A_{prop})}{N_{prop}(\pi,A_{prop})}\\
        &\quad \text{ if } N_{prop}(\pi,A_{prop}) > 0
        \end{split}
    \end{equation}
\end{small}

For a set of plans $\Pi$, we average over plans:

\begin{small}
    \begin{equation}
        AcceptRate(\Pi) = \frac{1}{|\Pi|} \sum_{\pi \in \Pi} AcceptRate(\pi,A_{prop})
    \end{equation}
\end{small}

With $|\Pi|$ denoting the cardinality (number of plans) of the set $\Pi$. This indicator is operational: it offers a simple, plan-computable proxy that captures only one facet of agency. It reflects the player's freedom to act, but not their emotional experience nor the actual impact of their choices on the narrative, which is left for future work.

\subsection{Author Intent}
\subsubsection{Definition}

Authorial intention can be understood as the extent to which the human author anticipates possible narrative experiences or allows the system to adapt creatively to the user \cite{riedlInteractiveNarrativeIntelligent2013}. Due to the interactivity offered to the player and the player's intention, the author's intention may not be respected.

\subsubsection{Formalism}
Whether representing the intention of the author or the player, this intention is commonly represented as a plan (or set of plans) \cite{cardona-riveraSymbolicPlanRecognition2015,riedlInteractiveNarrativeIntelligent2013} that the actor wishes to follow. We then measure this intention in two ways.

(1) By passing through the islands representing key points in the story. In our system $\mathfrak{S}$, we add $m$ islands, denoted $I = \{ I_1, I_2, \ldots, I_m \}$, where $I_n \subseteq S$, $I_n \cap I_k = \emptyset \text{ for } n \neq k$ and $I_n \cap (\{ s_{init}\} \cup S_{end}) = \emptyset$. We wish to determine, for a player's plan $\pi$, the number $m'$ of islands passed through.

(2) By comparing $\pi$ with the author's plans, since an EM may create branches that pass through the islands yet stray from the author's intention. A function $\kappa(\Pi_{Author},\pi) \in [0,1]$ measures how closely $\pi$ aligns with the set $\Pi_{Author}$ of plans the author considers in line with their intention (0: total divergence, 1: perfect conformity).

\subsubsection{Associated indicators}

The author's intention can be represented as intermediate steps that must occur in our story \cite{zhuAuthorIntentionBased2012}. Alignment with the author's intention can be measured by counting the number of constraints satisfied within the generated plan \cite{porteousApplyingPlanningInteractive2010}. In our formalism, these steps are the islands. For a plan $\pi$, we denote $m'(\pi)$ as the number of islands actually crossed (at least one occurrence of a state belonging to the island), and we define the island compliance rate :

\begin{equation}
IslandsRate(\pi) = \frac{m'(\pi)}{m}
\end{equation}

For a set of plans $\Pi$, we calculate an overall measure using the average :

\begin{small}
    \begin{equation}
        IslandsRate(\Pi) = \frac{1}{|\Pi|} \sum_{\pi \in \Pi} IslandsRate(\pi)
    \end{equation}
\end{small}

\subsection{Robustness}
\subsubsection{Definition}

Robustness corresponds to the system's ability to handle situations where the player's actions can block the narrative experience from continuing \cite{riedlRobustAuthorableMultiplayer2011}. A slightly different definition relates robustness to the system's ability to handle situations in which the player exceeds predefined limits \cite{luoReviewInteractiveNarrative2015}. We define these limits as exits from the story space (i.e. the space of possible narrative trajectories) constrained by the author. This issue is commonly described as the boundary problem \cite{magerkoComparativeAnalysisStory2007}.

\subsubsection{Formalism}

We define robustness as the EM's ability to manage transitions (actions or events) that could lead to a problematic state $s \in S_{prob} \text{ with } \forall s \in S_{prob}, \forall t \in T, (s,t) \notin dom(\gamma)$ and $S_{prob} \cap S_{goal} = \emptyset$ (i.e. the limits of the space of possible narrative trajectories anticipated by the author).

\subsubsection{Associated indicators}

Robustness is often achieved through repair mechanisms. When the player's behaviour is not anticipated, the system generates alternative branches to return to the story space \cite{youngArchitectureIntegratingPlanbased2004a}. Other approaches focus on the finishability \cite{wareMultiAgentNarrativeExperience2019}, where the objective is to eliminate problematic transitions. We measure the EM's ability to avoid problematic states directly, through the sequence of states it traverses.

We distinguish completed plans (ending in a state $s_{end} \in S_{goal}$) from failed plans (ending in a state $s_{end} \in S_{prob}$), in which the system could not steer the story towards one of the objectives. A low completion rate may suggest insufficient robustness. The completion rate is the share of completed plans:

\begin{equation}
CompletionRate(\Pi) =
\frac{
\bigl|\{\, \pi \in \Pi \mid s_{end} \in S_{goal} \,\}\bigr|
}{
|\Pi|
}
\end{equation}

\section{Case study}
This case study illustrates the framework on the narrative paradox: keeping the scenario fixed and varying only the EM policy, we compare how management strategies affect the balance between properties.

\subsection{Presentation of the design used}

We use a story inspired by \textit{Little Red Riding Hood}, already employed in work on the formal modelling of INS \cite{clercModelingInteractiveNarrative}. It features Little Red Riding Hood, her grandmother, the wolf, and the hunter, whose role the player takes on: the objective is to intervene at the right moment to save the characters threatened by the wolf. The narrative is expected to pass through one key stage (the \textit{island}): the hunter kills the wolf while Little Red Riding Hood or her grandmother is still in its belly. Killing the wolf too early skips this stage.

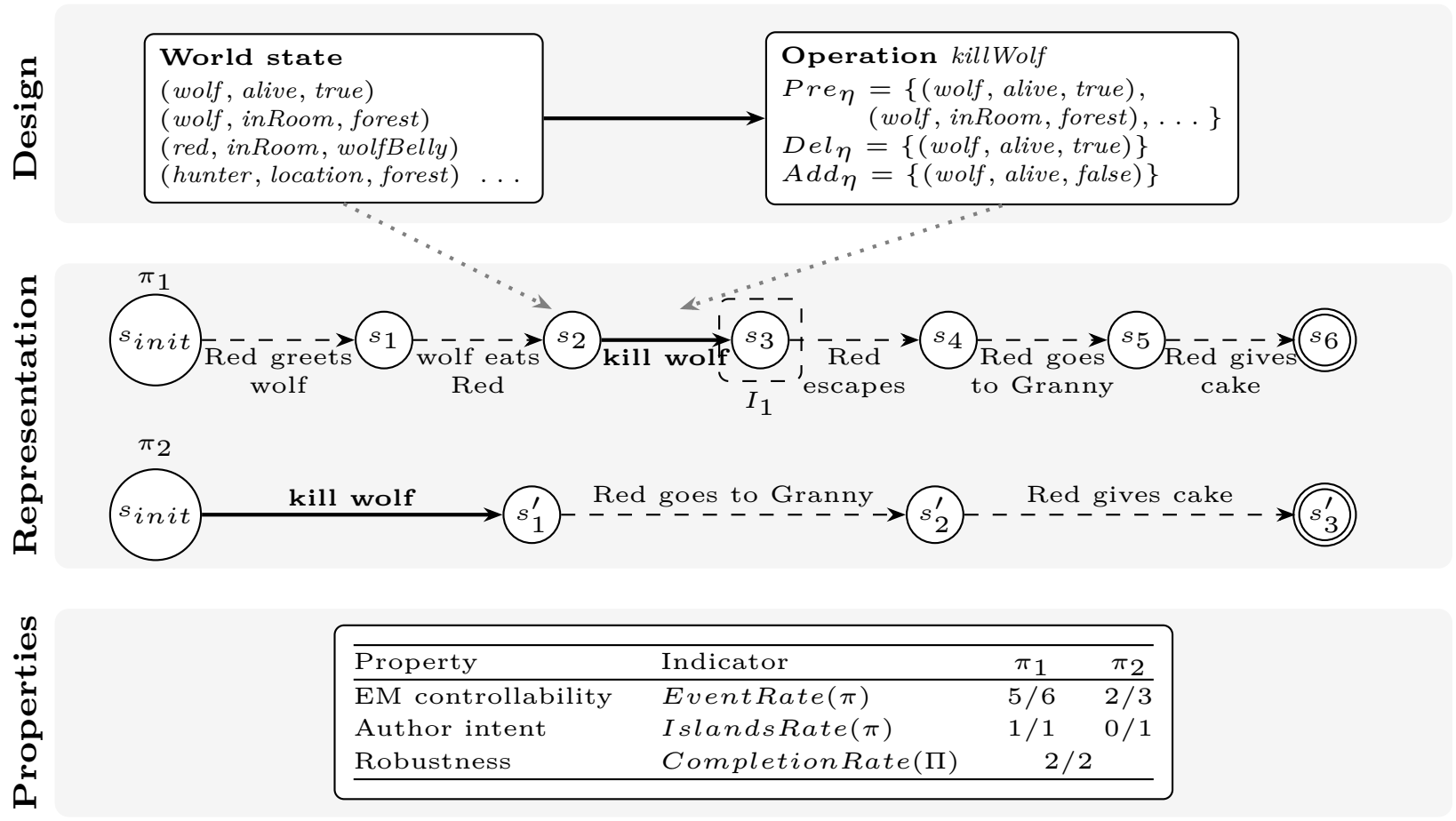
\begin{figure*}[t]
\centering
\begin{tikzpicture}[
  font=\scriptsize,
  box/.style  = {draw, rounded corners=2pt, align=left, inner sep=3pt,
                 fill=white, font=\tiny},
  st/.style   = {draw, circle, minimum size=12pt, inner sep=0pt,
                 fill=white, font=\tiny},
  goal/.style = {st, double, double distance=0.8pt},
  init/.style = {draw, circle, inner sep=1pt, minimum size=12pt,
                 fill=white, font=\tiny},
  act/.style  = {-{Stealth[length=4pt]}, thick},
  evt/.style  = {-{Stealth[length=4pt]}, dashed},
  map/.style  = {-{Stealth[length=4pt]}, dotted, gray, thick},
  lbl/.style  = {font=\tiny, align=center, inner sep=1.5pt},
  tbl/.style  = {draw, rounded corners=2pt, fill=white, inner sep=4pt,
                 font=\tiny},
  band/.style = {rotate=90, font=\scriptsize\bfseries}
]

%%%%%%%%%%%% bands %%%%%%%%%%%%
\fill[black!4, rounded corners=3pt] (0,3.62) rectangle (10.4,5.25);
\fill[black!4, rounded corners=3pt] (0,1.05) rectangle (10.4,3.32);
\fill[black!4, rounded corners=3pt] (0,-0.80) rectangle (10.4,0.75);
\node[band] at (-0.20,4.40) {Design};
\node[band] at (-0.20,2.19) {Representation};
\node[band] at (-0.20,-0.02) {Properties};

%%%%%%%%%%%% band 1 : design %%%%%%%%%%%%
\node[box] (ws) at (2.15,4.40) {%
  \textbf{World state}\\[1pt]
  $(\mathit{wolf},\mathit{alive},\mathit{true})$\\
  $(\mathit{wolf},\mathit{inRoom},\mathit{forest})$\\
  $(\mathit{red},\mathit{inRoom},\mathit{wolfBelly})$\\
  $(\mathit{hunter},\mathit{location},\mathit{forest})\ \dots$};

\node[box] (op) at (7.05,4.40) {%
  \textbf{Operation $\mathit{killWolf}$}\\[1pt]
  $Pre_\eta=\{(\mathit{wolf},\mathit{alive},\mathit{true}),$\\
  $\qquad\ \ (\mathit{wolf},\mathit{inRoom},\mathit{forest}),\dots\}$\\
  $Del_\eta=\{(\mathit{wolf},\mathit{alive},\mathit{true})\}$\\
  $Add_\eta=\{(\mathit{wolf},\mathit{alive},\mathit{false})\}$};

\draw[act] (ws) -- node[lbl,above] {$ $} (op);

%%%%%%%%%%%% band 2 : plan pi_1 %%%%%%%%%%%%
\node[init] (s0) at (0.75,2.75) {$s_{init}$};
\node[st]   (s1) at (2.45,2.75) {$s_1$};
\node[st]   (s2) at (3.85,2.75) {$s_2$};
\node[st]   (s3) at (5.25,2.75) {$s_3$};
\node[st]   (s4) at (6.65,2.75) {$s_4$};
\node[st]   (s5) at (8.05,2.75) {$s_5$};
\node[goal] (s6) at (9.45,2.75) {$s_6$};

\draw[evt] (s0) -- node[lbl,below] {Red greets\\wolf} (s1);
\draw[evt] (s1) -- node[lbl,below] {wolf eats\\Red} (s2);
\draw[act] (s2) -- node[lbl,below] {\textbf{kill wolf}} (s3);
\draw[evt] (s3) -- node[lbl,below] {Red\\escapes} (s4);
\draw[evt] (s4) -- node[lbl,below] {Red goes\\to Granny} (s5);
\draw[evt] (s5) -- node[lbl,below] {Red gives\\cake} (s6);

\node[draw, dashed, rounded corners=2pt, fit=(s3), inner sep=2.5pt] (isl) {};
\node[lbl] at (5.25,2.28) {$I_1$};
\node[lbl] at (0.75,3.19) {$\pi_1$};

%%%%%%%%%%%% band 2 : plan pi_2 %%%%%%%%%%%%
\node[init] (t0) at (0.75,1.45) {$s_{init}$};
\node[st]   (t1) at (3.55,1.45) {$s'_1$};
\node[st]   (t2) at (6.55,1.45) {$s'_2$};
\node[goal] (t3) at (9.45,1.45) {$s'_3$};

\draw[act] (t0) -- node[lbl,above] {\textbf{kill wolf}} (t1);
\draw[evt] (t1) -- node[lbl,above] {Red goes to Granny} (t2);
\draw[evt] (t2) -- node[lbl,above] {Red gives cake} (t3);
\node[lbl] at (0.75,1.95) {$\pi_2$};

%%%%%%%%%%%% band 3 : properties %%%%%%%%%%%%
\node[tbl] (pr) at (5.20,-0.02) {%
  \setlength{\tabcolsep}{5pt}%
  \renewcommand{\arraystretch}{1.15}%
  \begin{tabular}{@{}llcc@{}}
  \hline
  Property & Indicator & $\pi_1$ & $\pi_2$\\
  \hline
  EM controllability & $EventRate(\pi)$      & $5/6$ & $2/3$\\
  Author intent      & $IslandsRate(\pi)$    & $1/1$ & $0/1$\\
  Robustness         & $CompletionRate(\Pi)$ & \multicolumn{2}{c}{$2/2$}\\
  \hline
  \end{tabular}};

%%%%%%%%%%%% mapping arrows %%%%%%%%%%%%
\draw[map] (ws.south) -- (3.70,2.98);
\draw[map] (op.south) -- (4.65,2.98);

\end{tikzpicture}
\caption{The three layers of the framework, instantiated on the Little Red Riding Hood scenario. Design: a world state and the operation the player proposes. Representation: two plans produced from the initial state, with player actions as solid arrows, system events as dashed arrows, goal states as double circles, and the island $I_1$. Properties: the indicators computed from these plans. Both plans reach a goal state, but only $\pi_1$ crosses the island.}
\label{fig:layers}
\end{figure*}

Fig.~\ref{fig:layers} illustrates the three layers of the framework on this scenario: the design and its mapping onto the formal representation, detailed below, and the indicators computed from the resulting plans.

\subsubsection{World state \& Operations}

The design used for our INS is inspired by automated planning. In this approach, the system has, at any given time, an explicit description of the state of the world, and its evolution is described by operations composed of preconditions and effects. We chose this design because it makes it possible to describe what is true in the narrative world, what can be modified, and under what conditions these modifications may take place.

We represent the state of the world as a set of symbolic facts, expressed in the form of subject--predicate--object triples, in a way similar to the representations used in ontologies. A narrative fact is a triple $\tau = (u, p, o)$, where the subject $u$ is an entity of the narrative world, the predicate $p$ a relation or a property, and the object $o$ either another entity or a literal value. For instance, $(\mathit{wolf}, \mathit{inRoom}, \mathit{forest})$ locates the wolf, while $(\mathit{wolf}, \mathit{alive}, \mathit{true})$ relates it to a literal value. With $\mathcal{U}$ the set of entities, $\mathcal{P}$ the set of predicates and $\mathcal{L}$ the set of literal values, the set of possible narrative facts is $\mathcal{F} \subseteq \mathcal{U} \times \mathcal{P} \times (\mathcal{U} \cup \mathcal{L})$.

A world state $w_i \in W$, with $w_i \subseteq \mathcal{F}$, is a snapshot of the narrative world: the set of facts true at time $i$. The world evolves through narrative operations $\eta = (Pre_\eta, Add_\eta, Del_\eta)$, each modifying the world state: $Pre_\eta \subseteq \mathcal{F}$ is the set of facts that must be true for $\eta$ to be applied, and $Add_\eta, Del_\eta \subseteq \mathcal{F}$ are the facts it adds and removes. A narrative operation $\eta$ is applicable in a world state $w_i$ if its preconditions are true, i.e. $Pre_\eta \subseteq w_i$, and applying it produces the new world state $w_{i+1} = Result(w_i,\eta) = (w_i \backslash Del_\eta) \cup Add_\eta$.

A narrative operation transforms a snapshot of the narrative world into a new snapshot, by removing the facts that are no longer true and adding the facts produced by the action or event. Fig.~\ref{fig:layers} shows an example, the player's action $\mathit{killWolf}$.

\subsubsection{Experience Manager \& Player}

The set $\mathcal{H}$ of narrative operations is split into the actions $\mathcal{A}$ that the player may propose to the system and the events $\mathcal{E}$ controlled by the system: $\mathcal{H} = \mathcal{A} \cup \mathcal{E}$, with $\mathcal{A} \cap \mathcal{E} = \emptyset$.

At each time step $i$, the player does not access $w_i$ directly: they receive a textual observation, i.e. the initial text or the text of the last applied operation. From it, they may propose an action $\alpha_i \in \mathcal{A} \cup \{ \perp \}$, which the EM may accept, ignore, or replace with another narrative operation.

The EM then receives the current world state $w_i$ and the proposal $\alpha_i$, and selects the narrative operation to apply. This decision is represented by a policy function:

$$\nu_{EM} : W \times (\mathcal{A} \cup \{ \perp \} ) \rightarrow \mathcal{H}$$

We define the set of operations applicable in $w_i$ as:

$$\mathcal{H}(w_i)=\{\eta \in \mathcal{H} \mid Pre_\eta \subseteq w_i\}$$

The function $\nu_{EM}$ selects a narrative operation $\eta_i \in \mathcal{H}(w_i)$:

$$\eta_i = \nu_{EM}(w_i,\alpha_i)$$

If the EM accepts the player's proposal, then $\eta_i = \alpha_i$. The next world state is then obtained as follows:

$$w_{i+1}=Result(w_i,\eta_i)$$

Narrative goals are also sets of facts: the story is completed when all the facts of the goal $\mathcal{G} \subseteq \mathcal{F}$ are true, i.e. $\mathcal{G} \subseteq w_i$. Here, $\mathcal{G} = \{(granny, inInventory, cake)\}$.

\subsubsection{Mapping from the design to the representation}

This design is then mapped onto our formal representation:

$$\mathfrak{S} = (S, T , \gamma, s_{\text{init}}, S_{\text{goal}})$$

Since a world state is a set of facts true at a given time, each world state $w_i$ directly becomes a state $s_i \in S$ of the formal representation:

$$w_i \rightarrow s_i$$

The initial state is the initial world state, $s_{init} = w_0$, and the states are the world states reachable from $w_0$ by applying finite sequences of narrative operations from $\mathcal{H}$:

$$S= Reach(w_0,\mathcal{H})$$

Actions and events then become action-type and event-type transitions, respectively:

$$T = A \cup E \quad \text{with} \quad A=\mathcal{A}, \quad E=\mathcal{E}, \quad T=\mathcal{H}$$

The transition function $\gamma$ applies $Result$ whenever $\eta$ is applicable in $s$:

$$
    \gamma(s,\eta) = \left\{
          \begin{array}{ll}
            Result(s,\eta) & \quad \text{if} \quad Pre_{\eta} \subseteq s \\
            \text{not defined} & \quad \text{otherwise}
          \end{array}
        \right.
$$

Finally, a state $s \in S$ belongs to $S_{goal}$ if it satisfies the goal facts:

$$S_{\text{goal}} = \{s \in S \mid \mathcal{G} \subseteq s\}$$

\subsection{Simulation}

In order to carry out our functional analysis, we implemented $m = 3$ EMs with distinct story orchestration strategies. We ran $n = 100$ simulations for each EM, as described in pseudo-algorithm \ref{alg:multi_em_simulations}. We used an LLM\footnote{We use gpt-5-mini (with temperature set to $1$) both for the simulated player and for the generative behaviour of the EMs (action matching and dynamic action generation).} to simulate the behaviour of a human player, who may propose (or not propose) an action at each time step. We stopped the simulations once the running average of each indicator varied by less than $\varepsilon = 0.005$ over successive runs, which occurred at around $100$ simulations per EM.

\begin{algorithm}
    \caption{Simulate $n$ plans for $m$ EMs}
    \label{alg:multi_em_simulations}
    \textbf{Input}: System $\mathfrak{S}$\\
    \textbf{Parameter}: Number $n$ of simulations, List of $m$ EM $\{EM_j\}_{j=1..m}$\\
    \textbf{Output}: Set of plans $\Pi_{EM} = \{ \pi_{j,i} \}$ (a plan $\pi$ for EM $j$ and simulation $i$)
    \begin{algorithmic}[1]
        \FOR{$j \leftarrow 1$ \TO $m$}
            \FOR{$i \leftarrow 1$ \TO $n$}
                \STATE $s \leftarrow s_{init}$ ; $\pi \leftarrow [\,]$
                \WHILE{$s \notin S_{end}$}
                    \STATE Player proposes an action $a_{prop} \in \{\, a \in A \mid (s,a) \in dom(\gamma)\,\}\ \cup\ \{\perp\}$
                    \STATE $t_{select} \leftarrow \mu_{EM_j}(s,a_{prop})$
                    \STATE $s' \leftarrow \gamma(s,t_{select})$.
                    \STATE Add $s$ to $\pi$
                    \STATE $s \leftarrow s'$
                \ENDWHILE
                \STATE Add $s$ to $\pi$
                \STATE $\pi_{j,i} \leftarrow \pi$
            \ENDFOR
        \ENDFOR
        \STATE \textbf{return} $\Pi_{EM}$
    \end{algorithmic}
\end{algorithm}

We distinguish these EMs according to how they handle the player's proposals and according to the degree of control they exert over the evolution of the narrative.

The first manager (EM1) corresponds to a permissive and generative strategy. At each time step, when the player proposes an action in natural language, the EM first attempts to determine whether this action corresponds to an action already defined in the narrative design. If a match is found, the corresponding scripted action is selected and applied. However, if no existing action matches the player's proposal, the EM dynamically generates a new action from this proposal. This action is then integrated into the system as a new narrative operation, with its preconditions and effects, and may then be applied to the current world state. When the player does not propose any action, the EM triggers an available event in order to move the narrative forward. This strategy maximises the system's openness to the player's proposals, at the cost of dynamically extending the narrative space.

The second manager (EM2) corresponds to a scripted strategy. Unlike the first, it does not dynamically add new actions to the system. When a player proposes an action, the EM only checks whether this proposal corresponds to an action already planned in the design. If this is the case, the action is accepted and may be applied. If no match is found, the player's proposal is ignored or rejected, and the EM selects another available narrative operation, generally an event, in order to continue the execution of the scenario. This strategy limits the player's intervention to the actions anticipated by the designer, which reduces the openness of the system but ensures that the evolution of the narrative remains within the intended narrative space.

The third manager (EM3) corresponds to a scripted strategy with robustness control. Like the second EM, it only accepts actions that are already defined in the narrative design. However, before applying an action proposed by the player, it evaluates the state that would be obtained after triggering it. If the action leads to a valid state, it is accepted and applied. Conversely, if its application leads to a problematic state, the EM blocks this action and selects another applicable transition, for example a narrative event that makes it possible to maintain the progression of the narrative. This strategy introduces an additional control mechanism : it preserves the scripted framework of the scenario while avoiding situations likely to block the narrative experience.

\subsection{Results and analysis}

\begin{figure*}[t]
    \centering
    \includegraphics[width=0.85\textwidth]{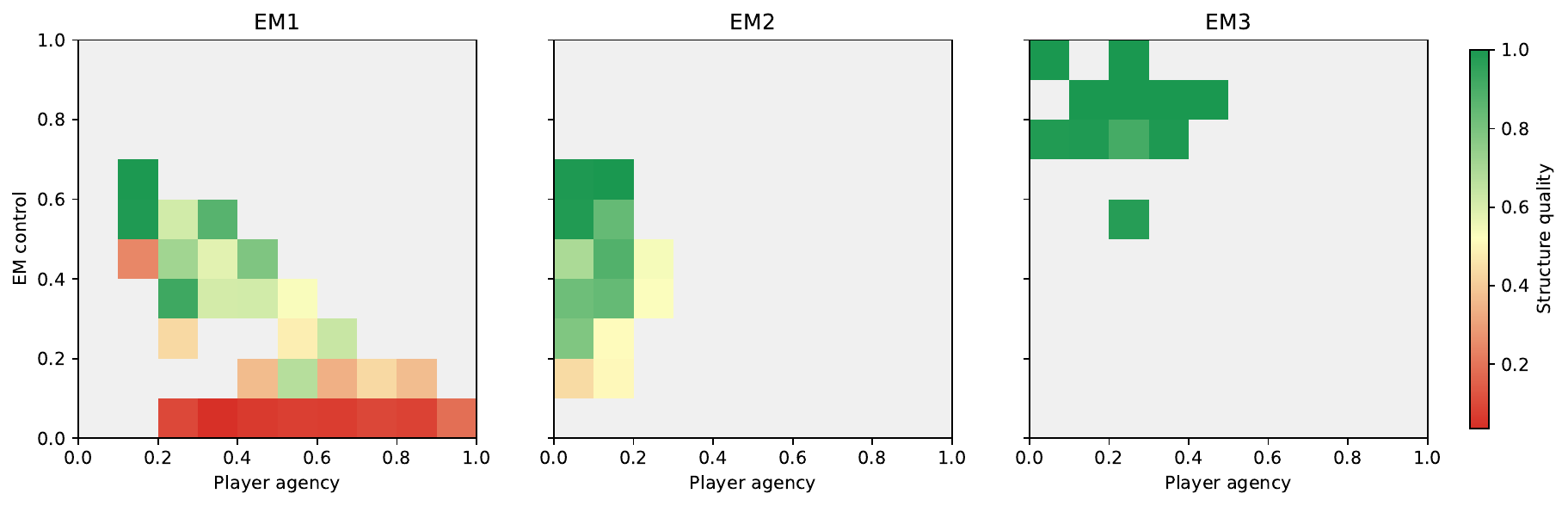}
    \caption{Heatmaps -- Narrative paradox trade-offs across EM policies.}
    \label{fig:heatmaps_em}
\end{figure*}

The quantitative results are reported in Table~\ref{tab:properties_results}, while Fig.~\ref{fig:heatmaps_em} provides a visual representation of the trade-offs between player agency, EM controllability, and narrative satisfaction.

The permissive and generative EM1 favours player agency, as shown by the highest $AcceptRate(\Pi)$ in Table~\ref{tab:properties_results}. By accepting or dynamically generating actions from the player's proposals, it gives more influence to the player. However, this openness may lead the story away from the authored narrative space and increase the risk of unstable or incomplete trajectories.

The scripted EM2 produces a more constrained behaviour. Since it only accepts actions already defined in the design, it reduces player agency but keeps the progression closer to the author's intended structure. This results in stronger narrative control, at the cost of limiting the player's influence on the story.

The scripted EM3 with robustness control introduces an additional filtering mechanism. By rejecting actions that would lead to problematic states, it preserves the scripted progression while avoiding situations that block the narrative. We do not report author intent and robustness for EM3 in Table~\ref{tab:properties_results}: both are maximal by construction, since this policy enforces passage through the islands and blocks transitions toward problematic states. This strategy improves robustness and author-intent alignment, but also reinforces the control exercised by the EM.

Fig.~\ref{fig:heatmaps_em} highlights the trade-off at the centre of the narrative paradox. High player agency tends to reduce control and may weaken the satisfaction of the generated structure, while stronger EM control improves robustness and author-intent alignment but restricts player freedom. The most interesting configurations are not located at the extremes, but in intermediate zones where the system preserves enough agency while maintaining a coherent and complete narrative progression.

Table~\ref{tab:properties_results} and Fig.~\ref{fig:heatmaps_em} thus help identify compromise zones between freedom, control, author intent, and robustness, which can guide the design of future EMs. Finally, each property is evaluated here through a single indicator. This keeps the analysis simple and reproducible, but additional indicators could be defined for each property to capture facets that a single measure leaves aside.

\begin{table}[t]
    \centering
    \caption{Properties results}\label{tab:properties_results}
    \footnotesize\setlength{\tabcolsep}{3.5pt}% HAL version: fit the column width
    \begin{tabular}{|l|l|c|c|c|}
        \hline
        Properties & Indicators & \textit{EM1} & \textit{EM2} & \textit{EM3} \\
        \hline
        Player agency & $AcceptRate(\Pi)$ & \textbf{0.49} & 0.11 & 0.19 \\
        EM controllability & $EventRate(\Pi)$ & 0.33 & 0.48 & \textbf{0.83} \\
        Author intent & $IslandsRate(\Pi)$ & 0.42 & 0.61 & \textbf{--} \\
        Robustness & $CompletionRate(\Pi)$ & 0.41 & 0.88 & \textbf{--} \\
        \hline
    \end{tabular}
\end{table}

\section{Discussion and perspectives}
% TODO, propriétés à définir dans le futur, pouvoir mettre en commun nos indicateurs afin de proposer un cadre d'évaluation commun. L'avantage du formalsime et de le partagé est qu'il va pouvoir être amélioré par la communauté de chercheurs. On a pas aussi parlé de l'aspect évaluation des observations NLP.

Our contribution is a shareable, reusable basis for modelling the properties involved in the narrative paradox, which the community can replicate and extend. The deliberate simplicity of the indicators, based on direct measurements from plans, prioritises transparency, reproducibility, and comparability across EMs. A first perspective is to consolidate the current properties and model new ones into a community-maintained catalogue.

The framework can also serve to prototype EM strategies: by simulating EM policies and analysing their effects on properties, it supports exploratory analysis and calibration during design. This use is complementary to user-based evaluations, which remain necessary to assess player experience but can benefit from prior filtering of EM behaviours.

Our study has limitations: we focused on simulation-based indicators with simplified EM policies, and did not yet evaluate generated narratives or natural-language interactions. Validating results with human players is therefore a key next step; integrating gameplay logs with participant-reported perceptions of structure and agency across repeated sessions \cite{fisherStructureAgencyImprovisation2025} offers a directly applicable model for this effort.

\section{Conclusion}
We proposed a methodology built on a formal INS model: clarifying properties from the state of the art, translating them into formal specifications, and making them assessable via indicators. This reduces ambiguities in property definitions and facilitates the comparison of INS, particularly around the narrative paradox.

We illustrated the framework through a case study on the narrative paradox. From simulated plans, we computed property indicators and used them to compare different EMs and visualise the trade-offs between player freedom and narrative control.

Future work will refine the evaluation of narrative satisfaction, better capture the effects of EM decisions on the player's experience, and turn the indicators into design criteria that help designers choose or configure an EM for specific narrative objectives. Finally, evaluating generated narratives and natural-language interactions is a key step, especially for emerging LLM-based storytelling.

\begingroup
\small
\subsubsection*{Acknowledgments.}
This work was supported by the CIFRE doctoral fellowship programme in collaboration with ALTEN and the Université de technologie de Compiègne.

\subsubsection*{Disclosure of Interests.}
The authors have no competing interests to declare that are relevant to the content of this article.
\par
\endgroup
%
% ---- Bibliography ----
%
\begingroup
\small
\bibliographystyle{splncs04}
\bibliography{references}
\endgroup

\end{document}